\documentclass{article}

\usepackage{graphicx}
\usepackage{amsmath, amssymb, amsfonts}
\usepackage{hyperref}
\usepackage{mathrsfs}
\usepackage{geometry}
\usepackage{algorithm}
\usepackage{algorithmic}
\usepackage{booktabs} 
\usepackage{multirow} 
\usepackage{caption}
\usepackage{subcaption} 
\usepackage{float}    
\usepackage{authblk}  

\title{Data-Efficient Meningioma Segmentation via Implicit Spatiotemporal Mixing and Sim2Real Semantic Injection\thanks{This study was supported by Shenzhen Basic Research Project (Natural Science Foundation) (Grant No.JCYJ20220530150416036)}\thanks{This study was a retrospective clinical study, and the study protocol was reviewed and approved by the ethics committee of our hospital, with the approval number: 20190910.}}

\author[1]{Yunhao Xu}
\author[3]{Fuquan Zong}
\author[3]{Yexuan Xing}
\author[1]{Chulong Zhang}
\author[1]{Guang Yang}
\author[2,1]{Shilong Yang}
\author[1]{Xiaokun Liang}
\author[3]{Juan Yu}

\affil[1]{Shenzhen Institute of Advanced Technology, Chinese Academy of Sciences, Shenzhen, China}
\affil[2]{Shenzhen University, Shenzhen, China}
\affil[3]{Department of Radiology, The First Affiliated Hospital of Shenzhen University, Health Science Center, Shenzhen Second People's Hospital, 3002 SunGangXi Road, 518035, Shenzhen, Guangdong, China}
\date{}

\begin{document}
\maketitle

\begin{abstract}
The performance of medical image segmentation is increasingly defined by the efficiency of data utilization rather than merely the volume of raw data. Accurate segmentation, particularly for complex pathologies like meningiomas, demands that models fully exploit the latent information within limited high-quality annotations. To maximize the value of existing datasets, we propose a novel dual-augmentation framework that synergistically integrates spatial manifold expansion and semantic object injection. Specifically, we leverage Implicit Neural Representations (INR) to model continuous velocity fields. Unlike previous methods, we perform linear mixing on the integrated deformation fields, enabling the efficient generation of anatomically plausible variations by interpolating within the deformation space. This approach allows for the extensive exploration of structural diversity from a small set of anchors. Furthermore, we introduce a Sim2Real lesion injection module. This module constructs a high-fidelity simulation domain by transplanting lesion textures into healthy anatomical backgrounds, effectively bridging the gap between synthetic augmentation and real-world pathology. Comprehensive experiments on a hybrid dataset demonstrate that our framework significantly enhances the data efficiency and robustness of state-of-the-art models, including nnU-Net and U-Mamba, offering a potent strategy for high-performance medical image analysis with limited annotation budgets.
\end{abstract}

\section{Introduction}

Meningiomas originate from the arachnoid cap cells of the meninges and represent the most common primary intracranial tumors in the central nervous system \cite{wiemels2010epidemiology}. Although the majority of meningiomas are benign, they are frequently associated with significant mass effects, compressing adjacent brain tissue, vasculature, and cranial nerves, which can lead to severe neurological deficits \cite{buerki2018overview}. In clinical workflows, multi-modal magnetic resonance imaging (MRI) provides complementary anatomical and pathological information—such as T1-weighted images depicting anatomical structures and T2/FLAIR sequences highlighting edema—serving as the gold standard for surgical planning and radiochemotherapy regimens \cite{Upadhyay2011BJR,Shukla2017CCO}. 

Recently, with the rapid advancement of deep learning, fully convolutional neural networks (e.g., nnU-Net \cite{Isensee2021nnUNet}) and emerging large-kernel or state-space models (e.g., U-Mamba) have achieved remarkable performance. However, the success of these increasingly complex, data-driven models is predicated on the availability of massive, high-quality pixel-level annotations. In the medical imaging domain, acquiring such datasets is hindered by the "annotation bottleneck": strict patient privacy regulations and the prohibitive cost of expert annotation create a disparity between the capability of modern models and the quantity of available labeled data \cite{Guan2021DASurvey}. This issue transforms the challenge from merely "obtaining data" to "efficiently utilizing limited data." Standard training sets often fail to cover the immense variability in tumor shape, location, and texture, causing models to suffer from a sharp decline in generalization when encountering unseen anatomical variations. Therefore, maximizing the utility of existing samples to simulate the true data distribution is crucial for training robust medical AI models.

Data augmentation—artificially expanding the diversity of training samples—is a key strategy to mitigate few-shot overfitting and enhance model robustness. Traditional augmentation methods primarily rely on affine transformations (e.g., rotation, scaling) and pixel-level perturbations (e.g., Gaussian noise, Gamma correction). While easy to implement, these methods perform linear interpolation only within the local neighborhood of existing data; they fail to generate samples with novel topological structures and thus cannot genuinely expand the data manifold \cite{Shorten2019AugmentSurvey}. To simulate non-linear soft tissue deformations, elastic deformation \cite{Krivov2017ElasticAug} has been widely adopted in medical imaging. However, due to the lack of physical constraints, randomly generated displacement fields are prone to producing anatomically implausible artifacts, such as ventricular collapse, cortical tearing, or tissue folding, which may mislead model feature learning. Recently, registration methods based on diffeomorphism (e.g., LDDMM \cite{Beg2005LDDMM}) have ensured topological fidelity by introducing fluid mechanics constraints. Notably, the advent of Implicit Neural Representations (INR) has enabled the construction of high-resolution, memory-efficient velocity fields in the continuous domain \cite{Sitzmann2020SIREN,Essakine2024INRSurvey}. Nevertheless, whether explicit elastic deformation or implicit diffeomorphism, these approaches essentially remain geometric resampling in the "spatial domain." While they enrich the anatomical morphology of normal brain tissue, they are constrained by physical laws and cannot generate new lesion semantics de novo, proving inadequate when addressing class imbalance or monotonic lesion morphology \cite{Dru2009SymLogDemons,Balakrishnan2019VoxelMorph}.

Another avenue for data expansion is semantic augmentation. Generative Adversarial Networks (GANs) \cite{Goodfellow2014GAN} and Diffusion Models \cite{Ho2020DDPM} have demonstrated the potential to generate high-fidelity medical images. However, in medical few-shot scenarios, these parameter-heavy models face severe training difficulties: extreme data scarcity often leads to discriminator overfitting or diffusion mode collapse. Generated samples are frequently simple memorizations or recombinations of training data, and worse, may produce "hallucinated" artifacts where pathological features do not match the anatomical background, lacking clinical credibility \cite{Goceri2023AugReview}. Conversely, "Copy-Paste" strategies represented by MixUp and CarveMix offer a non-parametric alternative by transplanting lesions from one sample to another \cite{Zhang2017Mixup,Zhang2023CarveMix,Ghiasi2021CopyPaste}. Although efficient, existing cut-and-paste methods often ignore tissue deformation caused by the tumor mass effect \cite{Clatz2005MassEffect}, and simple pixel blending struggles to handle the complex textural transitions between lesions and healthy tissue. This synthesis approach, lacking biomechanical consistency and textural continuity, induces significant domain shifts \cite{Kondrateva2021DomainShift}, forcing segmentation networks to learn incorrect boundary features and potentially compromising segmentation accuracy.

In summary, existing 3D medical image augmentation methods face a core dilemma: physics-driven deformation methods lack semantic creativity, while data-driven generative methods lack anatomical controllability and training stability. To address this challenge, this paper proposes the "Medical Image Augmentation Framework: Spatiotemporal Mixing and Semantic Injection," a dual-augmentation framework that integrates implicit fluid dynamics with explicit lesion injection. The core philosophy is to decouple and recombine "anatomical background diversity" and "lesion semantic richness." Specifically, we design two synergistic pathways: (1) Spatiotemporal Mixing: Utilizing Implicit Neural Representations (INR) to construct continuous Stationary Velocity Fields (SVF) in the latent space. By linearly weighting these velocity fields and solving the integration over time, we generate anatomical variants with strict diffeomorphic properties, thereby infinitely expanding the anatomical background without violating brain topology \cite{Dru2009SymLogDemons,Balakrishnan2019VoxelMorph}; (2) Semantic Injection: We propose a Sim2Real-driven lesion transplantation mechanism. This approach aims to minimize the domain shift between synthetic and real data by using a Distance Transform Map to guide the fusion of lesions into new anatomical backgrounds, simulating the infiltrative edge characteristics of meningiomas and eliminating synthesis artifacts \cite{Felzenszwalb2012DT}.

The main contributions of this study are summarized as follows:
\begin{enumerate}
    \item We propose a Spatiotemporal Mixing data augmentation strategy based on Implicit Neural Representations. Unlike traditional discrete augmentation, we utilize INR to construct a continuous anatomical deformation space. By performing mixing sampling in the velocity field domain rather than the pixel domain, we achieve anatomically enhanced generation with fluid mechanical constraints at a low computational cost, resolving the issue of topological disruption common in traditional elastic deformations.
    \item We introduce a Semantic Injection module. Addressing the difficulty of training generative models in few-shot scenarios, we introduce a training-free semantic synthesis scheme. By "planting" real lesion textures into deformed healthy brain tissues, we achieve synthetic data that combines anatomical structural diversity (from INR deformation) with high-fidelity lesion texture (from real sampling), mitigating the risk of generative hallucinations.
    \item We validate the framework on a hybrid dataset comprising private clinical data and public datasets. By applying Spatiotemporal Mixing to a small portion of labeled data and performing Semantic Injection simulations using a larger volume of tumor-free healthy data, we significantly improve segmentation performance.
\end{enumerate}

\section{Method}

Let $I \in \mathbb{R}^{D \times H \times W}$ denote the input image (assumed to be a single-channel 3D grayscale medical volume), where $D$, $H$, and $W$ represent depth, height, and width, respectively. The objective is to predict the semantic segmentation mask $M \in \{0, 1\}^{D \times H \times W}$. While generalized to $C$ classes, this study primarily focuses on the binary classification task (tumor versus background).

The datasets are defined as follows:
\begin{itemize}
    \item \textbf{Tumor Dataset} $\mathcal{D}_{tumor} = \{ (I_i, M_i) \}_{i=1}^{N}$, containing images with lesions and their corresponding annotations.
    \item \textbf{Normal Dataset} $\mathcal{D}_{norm} = \{ N_j \}_{j=1}^{M}$, consisting solely of normal brain images without lesion annotations.
\end{itemize}

\begin{figure}[H]
    \centering
    \includegraphics[width=0.95\linewidth]{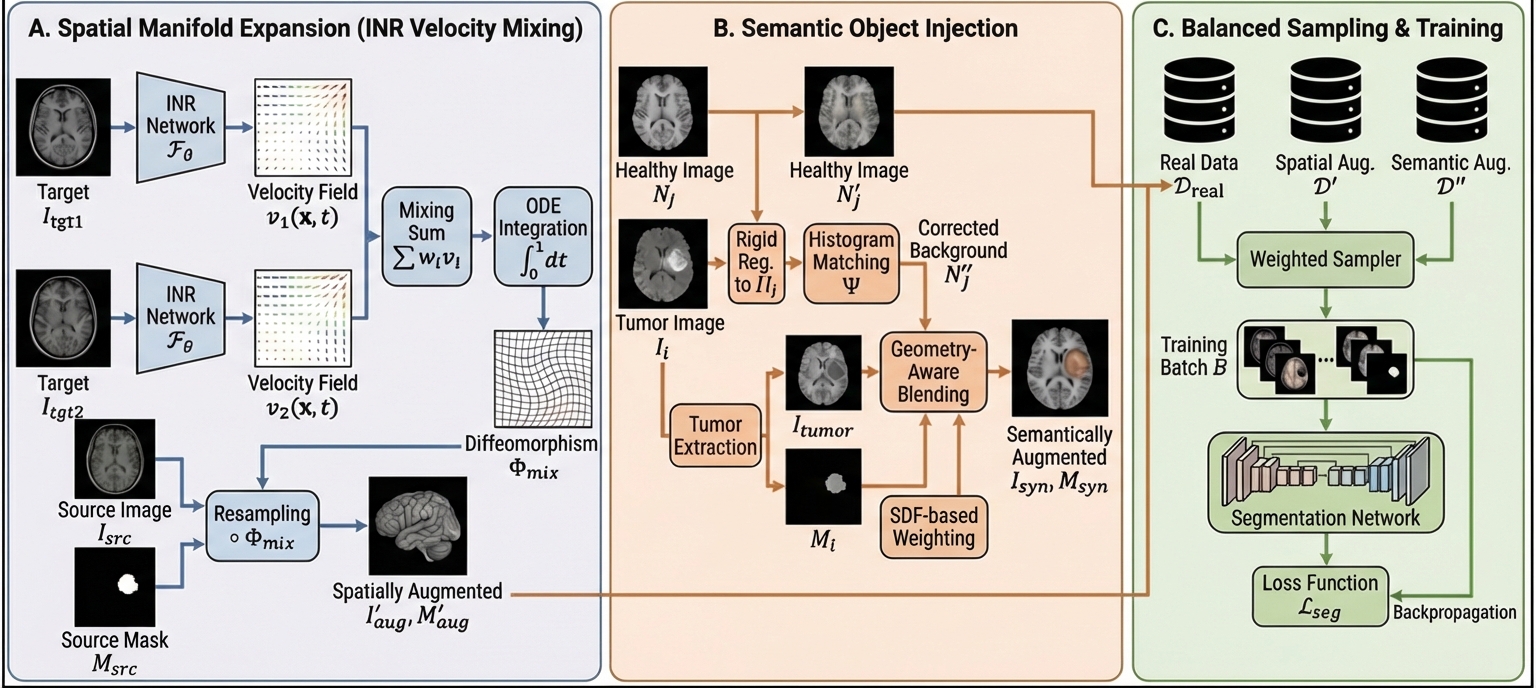}
    
    \caption{Overview of the proposed Dual-Augmentation Framework. It includes the INR-based spatial augmentation pathway, the semantic injection-based synthesis pathway, and the final weighted sampling training mechanism.}
    \label{fig:framework}
\end{figure}

The goal is to expand $\mathcal{D}_{tumor}$ via Spatial Augmentation and Semantic Augmentation to obtain the spatially augmented dataset $\mathcal{D}'$ and the semantically augmented dataset $\mathcal{D}''$, respectively. As illustrated in Fig. \ref{fig:framework}, our framework comprises two core modules: (A) an Implicit Neural Representation (INR)-based spatial manifold expansion module designed to generate augmented data with anatomical variations; and (B) a semantic injection module that utilizes data from healthy subjects to synthesize samples containing lesions. Ultimately, the real data and both types of synthetic data constitute a combined training pool, which is fed into the segmentation network via a weighted sampler.

\subsection{Spatial Augmentation via Implicit Neural Representations}

This module aims to generate anatomically plausible new morphologies by learning continuous registration fields between images and performing velocity field mixing.

\subsubsection{Preprocessing and Pairing}
All images are resampled to a unified spatial resolution $\Omega \subset \mathbb{R}^{D \times H \times W}$.
A pairing strategy is defined: for every $(I_{src}, M_{src}) \in \mathcal{D}_{tumor}$, two additional images are randomly sampled from $\mathcal{D}_{tumor}$ to serve as fixed targets, denoted as $I_{tgt1}$ and $I_{tgt2}$.
The objective is to compute the deformation field for the mapping $I_{src} \to I_{tgt}$.

\subsubsection{INR-based Diffeomorphic Sampling Field Modeling}
Based on diffeomorphic theory, this study utilizes Implicit Neural Representations (INR) to model the velocity field.
To resolve the conflict between the traditional forward flow (Source to Target) and the backward flow (Output Grid to Source) required for image interpolation, we directly define a backward velocity field $v(\mathbf{x}, t)$ within the target coordinate system.

\paragraph{ODE Formulation of Backward Sampling Mapping}
Let $t \in [0, 1]$ be a virtual time variable. We define $\phi(\mathbf{x}, t)$ as the sampling path at time $t$.
For any fixed point $\mathbf{x} \in \Omega$ on the target grid, its corresponding source image coordinate $\mathbf{y} = \phi(\mathbf{x}, 1)$ is obtained by integrating the following Ordinary Differential Equation (ODE):
\begin{equation}
    \frac{d\phi(\mathbf{x}, t)}{dt} = v(\phi(\mathbf{x}, t), t), \quad \phi(\mathbf{x}, 0) = \mathbf{x}
\end{equation}
Under this definition, $\phi(\mathbf{x}, 1)$ represents the sampling position corresponding to the target pixel $\mathbf{x}$ within the source image $I_{src}$.

\paragraph{Implicit Velocity Parameterization}
Unlike traditional methods that store velocity fields as discrete voxel grids, we employ a Coordinate-based Neural Network to approximate the velocity field function.

We define a spatiotemporal coordinate vector $\mathbf{c} = (\mathbf{x}, t) \in \mathbb{R}^4$. To mitigate the spectral bias of Multi-Layer Perceptrons (MLP) and enable better capture of high-frequency motion details, the coordinates are first subjected to positional encoding or Fourier Feature Mapping:
\begin{equation}
    \gamma(\mathbf{c}) = [\cos(2\pi \mathbf{B}\mathbf{c}), \sin(2\pi \mathbf{B}\mathbf{c})]^T
\end{equation}
where $\mathbf{B} \in \mathbb{R}^{L \times 4}$ is a frequency matrix sampled from a Gaussian distribution, and $L$ is the mapping dimension.

The velocity field is predicted by an MLP network $\mathcal{F}_\theta$ with parameters $\theta$:
\begin{equation}
    v(\mathbf{x}, t) = \mathcal{F}_\theta(\gamma(\mathbf{x}, t)) \in \mathbb{R}^3
\end{equation}
This implicit representation allows us to query velocity vectors at arbitrary continuous spatiotemporal coordinates, ensuring the mathematical continuity and resolution independence of the deformation field.

\paragraph{Discretized Integration}
During inference and training, the time domain is discretized into $K$ time steps. Let the time step size be $\Delta t = 1/K$.
By querying the implicit network $\mathcal{F}_\theta$, the deformation field $\phi_k$ at step $k$ can be updated via Euler integration approximation:
\begin{equation}
    \phi_{k+1}(\mathbf{x}) = \phi_k(\mathbf{x}) + \mathcal{F}_\theta(\gamma(\phi_k(\mathbf{x}), t_k)) \cdot \Delta t
\end{equation}
The final obtained deformation field is $\Phi = \phi_K$, which establishes the sampling mapping from the target image coordinate system to the source image coordinate system.
The corresponding displacement field is defined as $\mathbf{u}(\mathbf{x}) = \Phi(\mathbf{x}) - \mathbf{x}$.

\paragraph{Loss Function}
The loss function for training the network parameters $\theta$ consists of a similarity term and a regularization term:
\begin{equation}
    \mathcal{L} = \mathcal{L}_{sim}(I_{src} \circ \Phi, I_{tgt}) + \lambda \mathcal{L}_{reg}(v)
\end{equation}
where $\circ$ denotes the resampling of $I_{src}$ using the deformation field $\Phi$, and $\mathcal{L}_{reg}$ represents constraints on the velocity field gradients to ensure the smooth topological properties of the deformation.

\subsubsection{Velocity Field Mixing and Integration for Deformation Generation}

\paragraph{Velocity Field Mixing}
Assume the network has predicted two backward velocity fields $v_1(\mathbf{x}, t)$ and $v_2(\mathbf{x}, t)$ for $I_{tgt1}$ and $I_{tgt2}$, respectively.
We define the mixed velocity field $v_{mix}$ as:
\begin{equation}
    v_{mix}(\mathbf{x}, t) = w_1 \cdot v_1(\mathbf{x}, t) + w_2 \cdot v_2(\mathbf{x}, t)
\end{equation}
where $w_1, w_2 \in [0, 1]$ and $w_1 + w_2 = 1$.

\paragraph{Obtaining the Displacement Field}
The ODE is solved using the mixed velocity field $v_{mix}$ to obtain the final sampling mapping $\Phi_{mix}$. With a time step of $\Delta t = 1/K$:
\begin{align}
    \phi_{mix}^{(0)}(\mathbf{x}) &= \mathbf{x} \\
    \phi_{mix}^{(k+1)}(\mathbf{x}) &= \phi_{mix}^{(k)}(\mathbf{x}) + v_{mix}(\phi_{mix}^{(k)}(\mathbf{x}), t_k) \cdot \Delta t
\end{align}
The final sampling mapping is $\Phi_{mix}(\mathbf{x}) = \phi_{mix}^{(K)}(\mathbf{x})$.

\subsubsection{Application of Augmentation}
The source image and its label are spatially resampled using the derived mixed sampling mapping $\Phi_{mix}$:
\begin{align}
    I'_{aug}(\mathbf{x}) &= I_{src}(\Phi_{mix}(\mathbf{x})) \\
    M'_{aug}(\mathbf{x}) &= M_{src}(\Phi_{mix}(\mathbf{x}))
\end{align}
Here, $\Phi_{mix}(\mathbf{x})$ directly provides the sampling coordinates within the source image $I_{src}$.
Trilinear interpolation is applied to image $I$, while nearest-neighbor interpolation is used for label $M$. This process generates the spatially augmented dataset $\mathcal{D}'$.

\subsection{Semantic Augmentation}

This module aims to synthesize high-fidelity samples containing lesions by leveraging the rich pathological morphologies from $\mathcal{D}_{tumor}$ and the diverse healthy anatomical backgrounds from $\mathcal{D}_{norm}$ via a "Copy-Paste" strategy. To address issues such as anatomical misalignment, intensity inconsistency, and edge artifacts caused by direct transplantation, we propose a complete pipeline encompassing anatomical alignment, local intensity adaptation, and geometry-aware fusion.

\subsubsection{Anatomical Spatial Alignment}
Meningioma growth exhibits specific anatomical dependencies (typically adhering to the meninges). To ensure the anatomical plausibility of synthetic samples, we must unify the healthy background image $N_j$ and the lesion source image $I_i$ into the same anatomical coordinate system.

\paragraph{Rigid Registration}
Unlike simple geometric center alignment, we employ a rigid registration strategy based on Mutual Information (MI). We compute an affine transformation matrix $\mathbf{A} \in \mathbb{R}^{4 \times 4}$ such that the brain parenchyma structure of the healthy image $N_j$ aligns as closely as possible with that of $I_i$:
\begin{equation}
    \mathbf{A}^* = \arg\max_{\mathbf{A}} \text{MI}(I_i, \mathcal{T}(N_j, \mathbf{A}))
\end{equation}
Using $\mathbf{A}^*$, we resample $N_j$ to obtain the aligned background $N'_j$. This step ensures that the subsequently transplanted lesion remains intracranial on $N'_j$ and that its relative position (e.g., frontal or parietal lobe) remains invariant.

\subsubsection{Local Intensity Adaptation}
Due to gain differences across different scanners, direct fusion can lead to significant intensity discontinuities. To prevent global histogram matching from being distorted by high-signal tumor regions, we adopt a mask-constrained histogram matching approach.

Let $M_{brain}$ be the brain mask and $M_{tumor}$ be the lesion mask. We calculate the Cumulative Distribution Function (CDF) using only the statistics of the healthy parenchyma region $\Omega_{ref} = M_{brain} \setminus M_{tumor}$. We seek a mapping $\Psi$ such that the intensity distribution of $N'_j$ in the parenchyma region approximates the background distribution of $I_i$. The corrected background is denoted as $N''_j = \Psi(N'_j)$.

\subsubsection{Seamless Fusion Based on Distance Fields}
To eliminate the aliasing effects and unnatural texture discontinuities at the lesion boundary caused by the "cut-and-paste" operation, we propose a geometry-aware Alpha blending mechanism based on the Signed Distance Field (SDF).

\paragraph{Construction of Inward-Shrinking Mixing Weights}
Direct feathering often introduces invalid pixels from outside the lesion (e.g., the black background from the source image). We design an "inward mixing" strategy.
First, we compute the Signed Distance Field $SDF(\mathbf{x})$ of the lesion mask $M_i$, where the interior is positive. We define a transition bandwidth $\tau$ (typically set to 3-5 voxels). The Alpha weight map $\alpha(\mathbf{x})$ is constructed as follows:
\begin{equation}
    \alpha(\mathbf{x}) = \operatorname{clamp}\left( \frac{SDF(\mathbf{x})}{\tau}, 0, 1 \right)
\end{equation}
This weight map exhibits the following properties:
\begin{itemize}
    \item Core Region ($SDF > \tau$): $\alpha=1$, fully preserving the lesion texture $I_i$.
    \item Exterior Region ($SDF \le 0$): $\alpha=0$, fully preserving the background texture $N''_j$.
    \item Transition Layer ($0 < SDF \le \tau$): Linear interpolation.
\end{itemize}
This design ensures that the pixels from $I_i$ participating in the mixing always originate from the valid internal region of the lesion, completely preventing the introduction of background noise from the source image.

\paragraph{Simulation of Partial Volume Effect and Synthesis}
Real MRI scans exhibit the Partial Volume Effect (PVE) at tissue interfaces. To simulate this physical phenomenon and conceal splicing traces, we apply a weak low-pass filter to the background image at the boundaries prior to fusion.
We generate a boundary region mask $M_{border} = \mathbb{I}(0 < |SDF(\mathbf{x})| < \tau)$ and apply a Gaussian blur kernel $G_\sigma$ within this region:
\begin{equation}
    N'''_j(\mathbf{x}) = \begin{cases} 
    (N''_j * G_\sigma)(\mathbf{x}) & \text{if } \mathbf{x} \in M_{border} \\
    N''_j(\mathbf{x}) & \text{otherwise}
    \end{cases}
\end{equation}
The final synthetic image $I_{syn}$ is generated via weighted fusion:
\begin{equation}
    I_{syn}(\mathbf{x}) = \alpha(\mathbf{x}) \cdot I_i(\mathbf{x}) + (1 - \alpha(\mathbf{x})) \cdot N'''_j(\mathbf{x})
\end{equation}
The synthetic label directly inherits the original lesion mask $M_i$. This method preserves the core texture details of the lesion while achieving a smooth, natural transition into the new anatomical background.

\section{Experimental Settings}

\subsection{Datasets and Split}
\begin{itemize}
    \item \textbf{Training Set}: Meningioma data from Shenzhen Second People's Hospital, comprising 73 cases.
    \item \textbf{Testing Set}: Meningioma data from Shenzhen Second People's Hospital, comprising 20 cases (independently split, never involved in augmentation or training).
    \item \textbf{Auxiliary Dataset}: IXI healthy T1 MRI dataset, comprising 362 cases (used solely for generating synthetic data).
\end{itemize}

\subsection{Construction of Synthetic Data Pools and Sampling}
\subsubsection{Construction of Synthetic Data Pools}
\begin{itemize}
    \item \textbf{Spatial Augmentation Pool ($\mathcal{D}'$)}: For the 73 real cases in the training set, $K_{spatial}=20$ different deformation variants are randomly generated for each case. The mixing weights $w$ follow a $Beta(2, 2)$ distribution. In total, $73 \times 20 = 1460$ spatially augmented samples are generated.
    \item \textbf{Semantic Augmentation Pool ($\mathcal{D}''$)}: The 362 cases of normal IXI data are used as backgrounds. For each normal case, a lesion is randomly sampled from the 73 real lesions for injection, and this process is repeated $K_{semantic}=5$ times (varying the injection location and rotation angle each time). In total, $362 \times 5 \approx 1810$ semantically augmented samples are generated.
\end{itemize}
The final total volume of synthetic data is approximately 3270 cases, far exceeding the original 73 real cases.

\subsubsection{Online Balanced Sampling}
If mixed directly for training, the massive volume of synthetic data would overwhelm the distributional characteristics of the real data. Therefore, we employ an enforced proportional sampling strategy within each training batch.
Let the Batch Size be $B$. In each iteration:
\begin{itemize}
    \item Sample $B \times r_{real}$ samples from the real dataset $\mathcal{D}_{real}$.
    \item Sample $B \times (1 - r_{real})$ samples from the synthetic data pool $\mathcal{D}' \cup \mathcal{D}''$.
\end{itemize}
In the default setting, we set $r_{real} = 0.5$, meaning real data and augmented data are fed into the network at a 1:1 ratio. This ensures that the model learns the diversity of the augmented data while anchoring itself to the distribution of the real data.

\subsection{Benchmark Models and Comparison Methods}
We validate our framework on the following four State-of-the-Art (SOTA) models:
\begin{enumerate}
    \item \textbf{nnU-Net} \cite{Isensee2021nnUNet}: The current \textit{de facto} standard baseline for medical image segmentation.
    \item \textbf{U-Mamba}: A U-shaped network combining the Mamba state space model, excelling in long-sequence modeling.
    \item \textbf{Swin-UMamba}: A hybrid architecture merging the strengths of Swin Transformer and Mamba.
    \item \textbf{LKM-UNet}: A recent CNN architecture based on Large Kernel convolutions.
\end{enumerate}
All comparative experiments are divided into two groups: "Baseline" (using only 73 real cases) and "+ Ours" (using the complete augmentation strategy).

\subsection{Implementation Details}
Experiments are implemented based on the PyTorch and MONAI frameworks. All input images are resampled to $1.0 \times 1.0 \times 1.0$ mm and normalized to zero mean and unit variance. The training patch size is $96 \times 96 \times 96$. The optimizer is AdamW with an initial learning rate of $1e-4$ and weight decay of $1e-5$. A cosine annealing learning rate schedule is employed. The maximum training epochs are set to 1000, with a Batch Size of 4.

\section{Experimental Results}

\subsection{Qualitative Demonstration of Augmentation Effects}
\begin{figure}[H]
    \centering
    \includegraphics[width=0.95\linewidth]{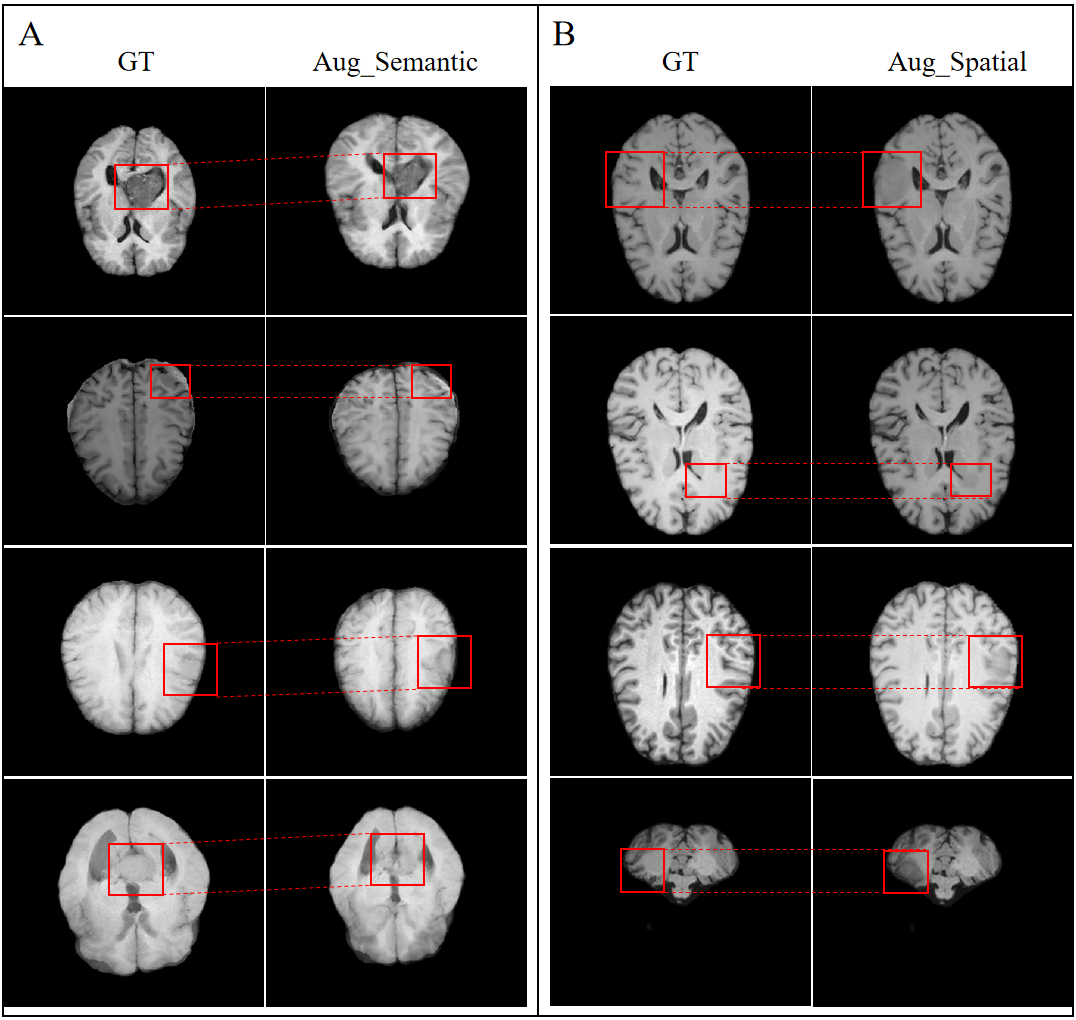}
    
    \caption{Qualitative demonstration of data augmentation effects. Figure 2(A) illustrates the Semantic Augmentation, which fills gaps in the pathological distribution by realistically embedding real lesion textures into healthy backgrounds. Figure 2(B) displays Spatial Augmentation based on INR velocity field mixing.}
    \label{fig:aug_qualitative}
\end{figure}

In this section, we first visually demonstrate the augmented data generated by our framework. Figure \ref{fig:aug_qualitative} illustrates the generation effects of spatial augmentation and semantic augmentation, respectively.

\subsection{Comparative Experiments with SOTA Methods}
Table \ref{tab:sota_comparison} presents the quantitative evaluation results on the independent test set comprising 20 cases. Experimental data indicate that the proposed dual-augmentation framework exhibits strong generalization capabilities, significantly enhancing the segmentation performance across various model architectures. Specifically, for the standard CNN architecture (nnU-Net), the Dice score improved by \textbf{9.96\%} relative to the baseline. Furthermore, on Transformer and Mamba-based models (e.g., Swin-UMamba), which are typically more sensitive to training data volume, our method successfully unlocks their long-range modeling potential. The performance improvement is even more pronounced, reaching up to \textbf{14.88\%}, with the final performance surpassing the CNN baseline. Additionally, the substantial overall reduction in the HD95 metric (e.g., Swin-UMamba dropping from 18.45 mm to \textbf{8.55 mm}) further confirms the precision of the augmented models in delineating complex boundaries.

\begin{table}[H]
    \centering
    \caption{Performance comparison (Mean) of different SOTA methods before and after applying our augmentation strategy. Data augmentation significantly improved performance across all architectures, particularly for data-sensitive Transformer/Mamba architectures.}
    \label{tab:sota_comparison}
    \renewcommand{\arraystretch}{1.3}
    \setlength{\tabcolsep}{8pt}
    \begin{tabular}{l|cc|cc}
        \toprule
        \multirow{2}{*}{\textbf{Method}} & \multicolumn{2}{c|}{\textbf{Dice Score (\%) $\uparrow$}} & \multicolumn{2}{c}{\textbf{HD95 (mm) $\downarrow$}} \\
        \cline{2-5} & \textbf{Baseline} & \textbf{Baseline + Ours} & \textbf{Baseline} & \textbf{Baseline + Ours} \\
        \hline
        nnU-Net       & 72.69 & \textbf{79.93} & 16.03 & \textbf{9.06} \\
        U-Mamba       & 71.25 & \textbf{80.15} & 17.82 & \textbf{8.92} \\
        Swin-UMamba   & 70.58 & \textbf{81.08} & 18.45 & \textbf{8.55} \\
        LKM-UNet      & 73.42 & \textbf{79.55} & 15.12 & \textbf{9.25} \\
        \bottomrule
    \end{tabular}
\end{table}

\subsection{Ablation Studies}
\begin{figure}[H]
    \centering
    \includegraphics[width=0.95\linewidth]{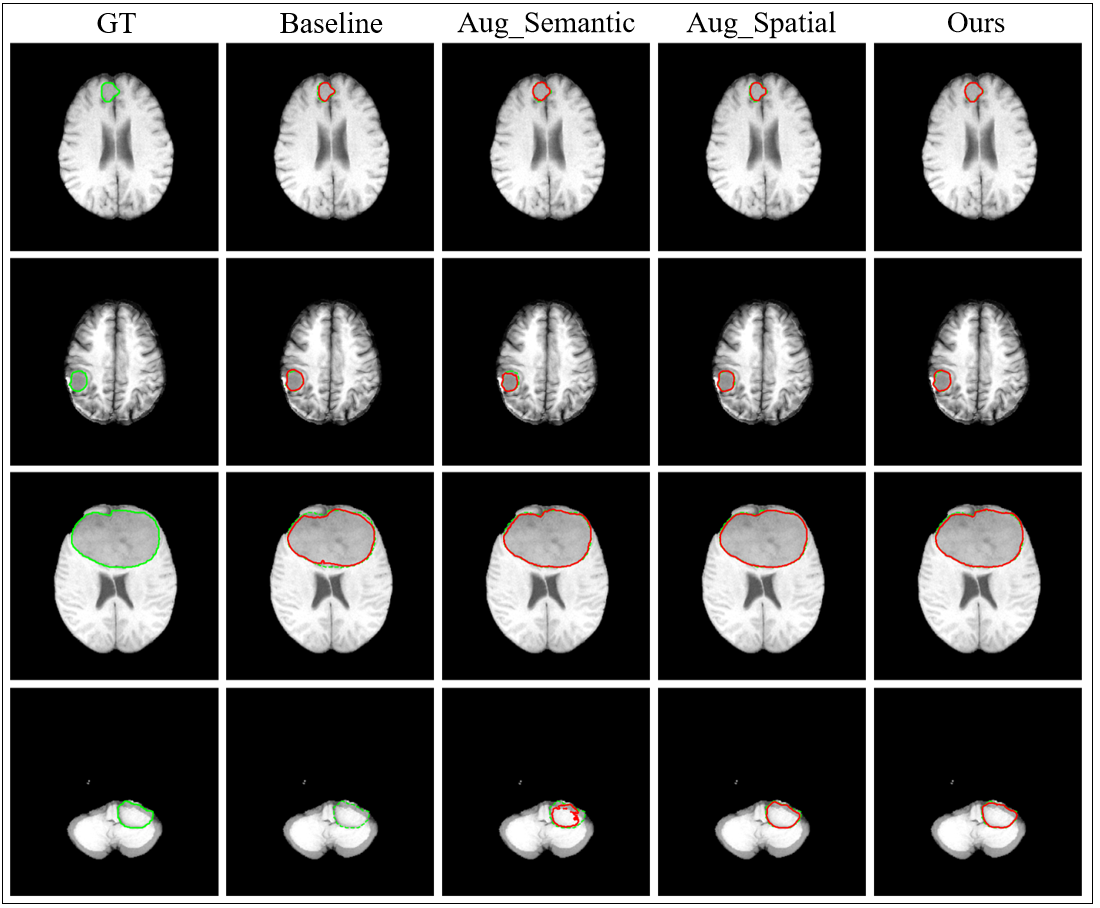}
    \caption{Visualization comparison of ablation studies. Each row represents a typical test sample. The green contour represents the Ground Truth (GT) annotated by experts, and the red contour represents the prediction results of each variant model.}
    \label{fig:ablation_visual}
\end{figure}

To validate the contribution of individual modules, we designed the following variants based on nnU-Net:
\begin{enumerate}
    \item \textbf{Baseline}: Real data only.
    \item \textbf{Spatial Only}: Real data + Spatial Augmentation Pool (1:1 sampling).
    \item \textbf{Semantic Only}: Real data + Semantic Augmentation Pool (1:1 sampling).
    \item \textbf{Full Pipeline}: Real data + Mixed Augmentation Pool (1:1 sampling).
\end{enumerate}
The quantitative results in Table \ref{tab:ablation} demonstrate that spatial augmentation and semantic augmentation improve model performance from the dimensions of "anatomical structural robustness" and "texture/background diversity," respectively. This conclusion is further corroborated by the visualization results in Fig. \ref{fig:ablation_visual}. Compared to the Baseline, the \textbf{Semantic Only} variant reduces missed detections (false negatives) by increasing texture diversity (as seen in the large lesion region in the third row). The \textbf{Spatial Only} variant, through geometric deformation training, results in smoother predicted boundaries that better adhere to anatomical rules. Finally, the \textbf{Full Pipeline} combines both strategies to yield a significant synergistic effect rather than interference, achieving precise capture of boundary details while maintaining internal consistency.

\begin{table}[h]
    \centering
    \caption{Ablation study results for each augmentation module (based on nnU-Net).}
    \label{tab:ablation}
    \renewcommand{\arraystretch}{1.2}
    \setlength{\tabcolsep}{12pt}
    \begin{tabular}{cc|cc}
        \toprule
        \textbf{Spatial Aug.} & \textbf{Semantic Aug.} & \textbf{Dice Score (\%) $\uparrow$} & \textbf{HD95 (mm) $\downarrow$} \\
        \hline
        - & - & 72.69 & 16.03 \\
        \checkmark & - & 75.29 & 9.69 \\
        - & \checkmark & 74.06 & 10.33 \\
        \checkmark & \checkmark & \textbf{79.93} & \textbf{9.06} \\
        \bottomrule
    \end{tabular}
\end{table}

\subsection{Evaluation of Synthetic Data Fidelity}
A core assumption of our Semantic Injection module is that the synthesized samples maintain sufficient fidelity to replace or supplement real data distributions. To explicitly evaluate this, we conducted an additional experiment comparing the performance of models trained on \textbf{purely synthesized data} versus \textbf{real data}.

Specifically, we designed three experimental settings based on the nnU-Net architecture:
\begin{itemize}
    \item \textbf{Real Data Only}: Training exclusively on the 73 original labeled cases.
    \item \textbf{Synthetic Data Only}: Training exclusively on the semantically augmented dataset $\mathcal{D}''$, derived from the healthy IXI dataset with injected lesions. Note that in this setting, the model never sees any real anatomical backgrounds from the target domain during training.
    \item \textbf{Mixed (Real + Synthetic)}: Training on a 1:1 mixture of both.
\end{itemize}

The results on the real test set are reported in Table \ref{tab:synthetic_fidelity}. It is observed that the model trained solely on synthetic data achieves a Dice score of 68.34\%. While this is slightly lower than the model trained on real data (72.69\%), the performance gap is remarkably narrow considering the synthetic model has never encountered the real patient backgrounds. This indicates that our injection strategy successfully preserves the critical semantic features of meningiomas and that the synthesized anatomical contexts are statistically close to the real distribution. Furthermore, when synthetic data is used to supplement real data, the performance significantly surpasses the baseline (74.06\%), confirming that the synthetic dataset provides complementary diversity rather than merely redundant noise.

\begin{table}[H]
    \centering
    \caption{Comparison of segmentation performance using purely synthetic data versus real data. The "Synthetic Data Only" setting demonstrates competitive performance, validating the high fidelity of our semantic injection strategy.}
    \label{tab:synthetic_fidelity}
    \renewcommand{\arraystretch}{1.3}
    \setlength{\tabcolsep}{10pt}
    \begin{tabular}{l|c|cc}
        \toprule
        \textbf{Training Data Source} & \textbf{Cases Used} & \textbf{Dice Score (\%) $\uparrow$} & \textbf{HD95 (mm) $\downarrow$} \\
        \hline
        Real Data Only (Baseline) & 73 & 72.69 & 16.03 \\
        Synthetic Data Only & 1810 & 68.34 & 18.21 \\
        Real + Synthetic Data & 73 + 1810 & \textbf{74.06} & \textbf{10.33} \\
        \bottomrule
    \end{tabular}
\end{table}

\section{Discussion}
The quantitative and qualitative results of this study strongly validate the effectiveness of decoupling anatomical variability from pathological semantics in medical image augmentation. The superiority of the proposed framework can be attributed to the synergistic interaction between its two core components. Unlike traditional elastic deformations that rely on random pixel-displacement fields, the INR-based spatial augmentation operates within the continuous velocity field domain. By performing linear mixing on velocity fields and solving the Ordinary Differential Equation (ODE) for integration, we ensure that the generated deformations strictly adhere to diffeomorphic properties. This mathematical constraint preserves the topological integrity of brain structures, effectively preventing anatomically implausible artifacts such as ventricular collapse or tissue tearing, which are common detrimental effects in standard data augmentation. Consequently, the model learns to recognize anatomical invariants under continuous geometric transformations, significantly enhancing its robustness to shape variations inherent in patient populations.

Furthermore, the differential improvements observed across model architectures highlight the specific value of our framework for modern, data-driven networks. While the CNN-based nnU-Net showed substantial gains, the performance leap was even more pronounced for the Mamba and Transformer-based architectures (e.g., Swin-UMamba). These advanced architectures typically lack the inherent inductive biases of CNNs (such as translation invariance) and thus require significantly larger datasets to learn effective representations. Our dual-augmentation strategy successfully bridges this gap by synthesizing a high-variance training distribution that mimics "Big Data," thereby preventing overfitting in these parameter-heavy models. This finding suggests that our framework is particularly well-suited for unlocking the potential of emerging foundation models in medical scenarios where labeled data remains scarce.

From a clinical perspective, the results offer promising implications for surgical planning and computer-aided diagnosis. The substantial reduction in the Hausdorff Distance (HD95) metric—dropping to as low as \textbf{8.55 mm}—indicates that the augmented models produce segmentation boundaries that are significantly tighter and geometrically more accurate. In neurosurgery, where the distinction between tumor margins and eloquent brain areas is critical, such precision is paramount for minimizing post-operative neurological deficits. Additionally, the Semantic Injection module demonstrates a cost-effective pathway for data utilization. By leveraging readily available, unlabelled healthy data (e.g., from public datasets like IXI) to synthesize pathological samples, we effectively bypass the strict privacy regulations and high annotation costs associated with acquiring real tumor datasets. This "synthetic mixing" approach provides a practical solution for hospitals to train robust models without necessitating massive internal annotation efforts.

Despite these promising results, several limitations warrant further investigation. First, the current semantic injection module relies on rigid registration and heuristic distance-based fusion. While this creates realistic textures, it does not fully simulate the non-linear mass effect (e.g., midline shift or compression of adjacent ventricles) typically caused by large meningiomas, potentially limiting the model's learning of global structural distortions. Second, compared to simple affine transformations, the solving of ODEs for INR-based deformation introduces a higher computational overhead during the training phase. Future work will focus on integrating biomechanical constraints into the injection process to simulate realistic mass effects and exploring distilled versions of INR to enhance training efficiency.

\section{Conclusion}
In this work, we proposed a novel dual-augmentation framework that addresses the annotation bottleneck in medical image segmentation by synergizing Implicit Neural Representations with semantic injection. By constructing continuous, topology-preserving anatomical variations and synthesizing high-fidelity lesion samples from healthy cohorts, our method effectively expands the training manifold without incurring additional annotation costs. Comprehensive experiments demonstrate that this approach significantly boosts the generalization capability of state-of-the-art models, particularly enabling data-hungry architectures like U-Mamba to achieve superior performance on small datasets. Ultimately, this framework offers a robust and data-efficient paradigm for high-performance medical image analysis, paving the way for more reliable clinical AI deployment in low-resource settings.

\bibliographystyle{IEEEtran}
\bibliography{refs}

@article{Upadhyay2011BJR,
  author  = {Upadhyay, Niranjan and Waldman, Adam D.},
  title   = {Conventional MRI evaluation of gliomas},
  journal = {The British Journal of Radiology},
  year    = {2011},
  volume  = {84},
  number  = {Suppl 2},
  pages   = {S107--S114},
  doi     = {10.1259/bjr/65711810},
  url     = {https://pmc.ncbi.nlm.nih.gov/articles/PMC3473894/}
}

@article{Shukla2017CCO,
  author  = {Shukla, Gaurav and Alexander, Gregory S. and Bakas, Spyridon and Nikam, Rahul and Talekar, Kiran and Palmer, Joshua D. and Shi, Wenyin},
  title   = {Advanced magnetic resonance imaging in glioblastoma: a review},
  journal = {Chinese Clinical Oncology},
  year    = {2017},
  volume  = {6},
  number  = {4},
  pages   = {40},
  doi     = {10.21037/cco.2017.06.28},
  url     = {https://cco.amegroups.org/article/view/15820/html}
}

@article{Isensee2021nnUNet,
  author  = {Isensee, Fabian and J{"a}ger, Paul F. and Kohl, Simon A. A. and Petersen, Jens and Maier-Hein, Klaus H.},
  title   = {nnU-Net: A self-configuring method for deep learning-based biomedical image segmentation},
  journal = {Nature Methods},
  year    = {2021},
  volume  = {18},
  pages   = {203--211},
  doi     = {10.1038/s41592-020-01008-z},
  url     = {https://www.nature.com/articles/s41592-020-01008-z}
}

@article{Guan2021DASurvey,
  author  = {Guan, Hao and Liu, Mingxia},
  title   = {Domain Adaptation for Medical Image Analysis: A Survey},
  journal = {IEEE Transactions on Biomedical Engineering},
  year    = {2021},
  volume  = {69},
  number  = {3},
  pages   = {1173--1185},
  url     = {https://pmc.ncbi.nlm.nih.gov/articles/PMC9011180/}
}

@article{Shorten2019AugmentSurvey,
  author  = {Shorten, Connor and Khoshgoftaar, Taghi M.},
  title   = {A survey on image data augmentation for deep learning},
  journal = {Journal of Big Data},
  year    = {2019},
  volume  = {6},
  number  = {1},
  doi     = {10.1186/s40537-019-0197-0},
  url     = {https://link.springer.com/article/10.1186/s40537-019-0197-0}
}

@inproceedings{Krivov2017ElasticAug,
  author    = {Krivov, Egor and Pisov, Maxim and Belyaev, Mikhail},
  title     = {MRI Augmentation via Elastic Registration for Brain Lesions Segmentation},
  booktitle = {International MICCAI BrainLesion Workshop},
  series    = {LNCS},
  year      = {2017},
  publisher = {Springer},
  doi       = {10.1007/978-3-319-75238-9_32},
  url       = {https://www.researchgate.net/publication/322007593}
}

@article{Beg2005LDDMM,
  author  = {Beg, Mirza Faisal and Miller, Michael I. and Trouv{'e}, Alain and Younes, Laurent},
  title   = {Computing Large Deformation Metric Mappings via Geodesic Flows of Diffeomorphisms},
  journal = {International Journal of Computer Vision},
  year    = {2005},
  volume  = {61},
  pages   = {139--157},
  doi     = {10.1023/B:VISI.0000043755.93987.aa},
  url     = {https://link.springer.com/article/10.1023/B:VISI.0000043755.93987.aa}
}

@inproceedings{Sitzmann2020SIREN,
  author    = {Sitzmann, Vincent and Martel, Julien N. P. and Bergman, Alexander W. and Lindell, David B. and Wetzstein, Gordon},
  title     = {Implicit Neural Representations with Periodic Activation Functions},
  booktitle = {Advances in Neural Information Processing Systems (NeurIPS)},
  year      = {2020},
  url       = {https://proceedings.neurips.cc/paper/2020/file/53c04118df112c13a8c34b38343b9c10-Paper.pdf}
}

@article{Essakine2024INRSurvey,
  author  = {Essakine, Amine and Cheng, Yang and Cheng, Chun-Wu and Zhang, Ling and Van Gool, Luc},
  title   = {Where Do We Stand with Implicit Neural Representations? A Technical and Performance Survey},
  journal = {arXiv},
  year    = {2024},
  eprint  = {2411.03688},
  url     = {https://arxiv.org/abs/2411.03688}
}

@techreport{Dru2009SymLogDemons,
  author      = {Dru, Florence and Vercauteren, Tom},
  title       = {An ITK Implementation of the Symmetric Log-Domain Diffeomorphic Demons Algorithm},
  institution = {Inria},
  year        = {2009},
  url         = {https://inria.hal.science/hal-00813744/}
}

@inproceedings{Goodfellow2014GAN,
  author    = {Goodfellow, Ian and Pouget-Abadie, Jean and Mirza, Mehdi and Xu, Bing and Warde-Farley, David and Ozair, Sherjil and Courville, Aaron and Bengio, Yoshua},
  title     = {Generative Adversarial Nets},
  booktitle = {Advances in Neural Information Processing Systems (NeurIPS)},
  year      = {2014},
  url       = {https://papers.nips.cc/paper_files/paper/2014/file/5ca3e9b122f61f8f06494c97b1afccf3-Paper.pdf}
}

@inproceedings{Ho2020DDPM,
  author    = {Ho, Jonathan and Jain, Ajay and Abbeel, Pieter},
  title     = {Denoising Diffusion Probabilistic Models},
  booktitle = {Advances in Neural Information Processing Systems (NeurIPS)},
  year      = {2020},
  url       = {https://arxiv.org/abs/2006.11239}
}

@article{Goceri2023AugReview,
  author  = {Goceri, Evgin},
  title   = {Medical image data augmentation: Techniques, comparisons and interpretations},
  journal = {Artificial Intelligence Review},
  year    = {2023},
  volume  = {56},
  pages   = {4123--4170},
  doi     = {10.1007/s10462-023-10453-z},
  url     = {https://link.springer.com/article/10.1007/s10462-023-10453-z}
}

@article{Zhang2017Mixup,
  author  = {Zhang, Hongyi and Cisse, Moustapha and Dauphin, Yann N. and Lopez-Paz, David},
  title   = {mixup: Beyond Empirical Risk Minimization},
  journal = {arXiv preprint arXiv:1710.09412},
  year    = {2017},
  url     = {https://arxiv.org/abs/1710.09412}
}

@article{Zhang2023CarveMix,
  author  = {Zhang, Xinyu and Liu, Chenghao and Ou, Na and Xu, Xiangzhen and Zhou, Zhiqiang and Duan, Yunyun and Xiong, Xianlong and You, Yuhong and Lv, Ziheng and Xu, You and Liu, Youand Bi, Chuangye},
  title   = {CarveMix: A simple data augmentation method for brain lesion segmentation},
  journal = {NeuroImage},
  year    = {2023},
  volume  = {268},
  pages   = {119858},
  doi     = {10.1016/j.neuroimage.2023.119858},
  url     = {https://www.sciencedirect.com/science/article/pii/S1053811923001878}
}

@inproceedings{Ghiasi2021CopyPaste,
  author    = {Ghiasi, Golnaz and Cui, Yin and Srinivas, Aravind and Qian, Rui and Lin, Tsung-Yi and Cubuk, Ekin D. and Le, Quoc V. and Zoph, Barret},
  title     = {Simple Copy-Paste Is a Strong Data Augmentation Method for Instance Segmentation},
  booktitle = {Proceedings of the IEEE/CVF Conference on Computer Vision and Pattern Recognition (CVPR)},
  year      = {2021},
  pages     = {2918--2928},
  url       = {https://openaccess.thecvf.com/content/CVPR2021/papers/Ghiasi_Simple_Copy-Paste_Is_a_Strong_Data_Augmentation_Method_for_Instance_CVPR_2021_paper.pdf}
}

@inproceedings{Kondrateva2021DomainShift,
  author    = {Kondrateva, Ekaterina and Pominova, Marina and Popova, Elena and et al.},
  title     = {Domain shift in computer vision models for MRI data analysis: an overview},
  booktitle = {Proceedings of SPIE, Thirteenth International Conference on Machine Vision (ICMV 2020)},
  year      = {2021},
  volume    = {11605},
  pages     = {116050H},
  doi       = {10.1117/12.2587872},
  url       = {https://arxiv.org/pdf/2010.07222}
}

@article{Clatz2005MassEffect,
  author  = {Clatz, Olivier and Sermesant, Maxime and Bondiau, Pierre-Yves and Delingette, Herv{'e} and Warfield, Simon K. and Malandain, Gr{'e}goire and Ayache, Nicholas},
  title   = {Realistic simulation of the 3-D growth of brain tumors in MR images coupling diffusion with biomechanical deformation},
  journal = {IEEE Transactions on Medical Imaging},
  year    = {2005},
  volume  = {24},
  number  = {10},
  pages   = {1334--1346},
  doi     = {10.1109/TMI.2005.857217},
  url     = {https://pmc.ncbi.nlm.nih.gov/articles/PMC2496876/}
}

@article{Felzenszwalb2012DT,
  author  = {Felzenszwalb, Pedro F. and Huttenlocher, Daniel P.},
  title   = {Distance transforms of sampled functions},
  journal = {Theory of Computing},
  year    = {2012},
  volume  = {8},
  number  = {1},
  pages   = {415--428},
  doi     = {10.4086/toc.2012.v008a019},
  url     = {http://theoryofcomputing.org/articles/v008a019/}
}

@article{Balakrishnan2019VoxelMorph,
  author  = {Balakrishnan, Guha and Zhao, Amy and Sabuncu, Mert R. and Guttag, John and Dalca, Adrian V.},
  title   = {VoxelMorph: A Learning Framework for Deformable Medical Image Registration},
  journal = {IEEE Transactions on Medical Imaging},
  year    = {2019},
  volume  = {38},
  number  = {8},
  pages   = {1788--1800},
  doi     = {10.1109/TMI.2019.2897538},
  url     = {https://ieeexplore.ieee.org/abstract/document/8633930/}
}

@article{wiemels2010epidemiology,
  title={Epidemiology and etiology of meningioma},
  author={Wiemels, Joseph and Wrensch, Margaret and Claus, Elizabeth B},
  journal={Journal of neuro-oncology},
  volume={99},
  number={3},
  pages={307--314},
  year={2010},
  publisher={Springer}
}

@article{buerki2018overview,
  title={An overview of meningiomas},
  author={Buerki, Robin A and Horbinski, Craig M and Kruser, Timothy and Horowitz, Peleg M and James, Charles David and Lukas, Rimas V},
  journal={Future oncology},
  volume={14},
  number={21},
  pages={2161--2177},
  year={2018},
  publisher={Taylor \& Francis}
}

\end{document}